\documentclass[10pt,twocolumn,letterpaper]{article}

\usepackage{3dv}
\usepackage{times}
\usepackage{epsfig}
\usepackage{graphicx}
\usepackage{amsmath}
\usepackage{amssymb}

\usepackage{textcomp}

\usepackage[caption = false]{subfig}
\usepackage{cite}
\usepackage{bm}
\usepackage{makecell}

\usepackage[pagebackref=true,breaklinks=true,letterpaper=true,colorlinks,bookmarks=false]{hyperref}

\DeclareMathOperator*{\argmin}{argmin}   

\usepackage{multirow}
\usepackage{xspace}
\makeatletter
\DeclareRobustCommand\onedot{\futurelet\@let@token\@onedot}
\def\@onedot{\ifx\@let@token.\else.\null\fi\xspace}
\def\eg{\emph{e.g}\onedot} 
\def\ie{\emph{i.e}\onedot} 
 
\def\etc{\emph{etc}\onedot}
\def\vs{\emph{vs}\onedot}
\def\wrt{\emph{w.r.t}\onedot}

\makeatother

\usepackage{lipsum}

\newcommand\blfootnote[1]{%
  \begingroup
  \renewcommand\thefootnote{}\footnote{#1}%
  \addtocounter{footnote}{-1}%
  \endgroup
}

\threedvfinalcopy 

\def\threedvPaperID{79} 

\ifthreedvfinal\fi
\begin{document}

\title{SMPLy Benchmarking 3D Human Pose Estimation in the Wild}

\author{Vincent Leroy$^{*}$
\and
Philippe Weinzaepfel$^{*}$
\and
Romain Br\'egier
\and
Hadrien Combaluzier \hspace{10mm}  Gr\'egory Rogez \vspace{1mm}\\
{NAVER LABS Europe}\\
{\tt\small firstname.lastname@naverlabs.com}
\footnotetext{ $^{*}$ indicates equal contribution.}
}

\maketitle
\thispagestyle{empty}

\begin{abstract}

Predicting 3D human pose from images has seen great recent improvements. Novel approaches that can even predict both pose and shape from a single input image have been introduced, often relying on a parametric model of the human body such as SMPL. While qualitative results for such methods are often shown for images captured in-the-wild, a proper benchmark in such conditions is still missing, as it is cumbersome to obtain ground-truth 3D poses elsewhere than in a motion capture room.
This paper presents a pipeline to easily produce and validate such a dataset with accurate ground-truth, with which we benchmark recent 3D human pose estimation methods in-the-wild. 
We make use of the recently introduced Mannequin Challenge dataset which contains in-the-wild videos of people frozen in action like statues and leverage the fact that people are static and the camera moving to accurately fit the SMPL model on the sequences. 
A total of 24,428 frames with registered body models are then selected from 567 scenes at almost no cost, using only online RGB videos.
We benchmark state-of-the-art SMPL-based human pose estimation methods on this dataset.
Our results highlight that  challenges  remain, in particular for difficult poses or for scenes where the persons are partially truncated or occluded.



\end{abstract}

\section{Introduction} 
\blfootnote{ $^{*}$ indicates equal contribution.}
Human pose estimation is an important computer vision problem with many possible applications in robotics, virtual/augmented reality or human-computer interactions.
Primarily driven by the availability (or absence) of training data, the problem has been originally tackled either as 2D pose estimation in-the-wild~\cite{mpii} or 3D pose estimation in either synthetic~\cite{unite} or constrained scenarios~\cite{h36,totalcapture}, \eg using Motion Capture (MoCap) rooms.
Recently, some works have managed to predict 3D poses for in-the-wild images, and impressively, various approaches can even predict the 3D human shape from a single image. 
Most of these methods rely on a parametric model of the human body such as SMPL~\cite{smpl}, and estimate the parameters that control pose and shape deformations of the model~\cite{hmr,unite,graphcmr,texturepose,dct,zanfir2018deep}. 
A few recent works make direct predictions of the human 3D shape without any parametric representation~\cite{moulding,zhu2019detailed,deephuman}. 
The task of human 3D pose and shape estimation is mostly evaluated on datasets captured in a constrained environment~\cite{humaneva,h36}, and only qualitative results are usually shown for a few images captured in the wild. 
Even if some efforts have been made to produce outdoor evaluation datasets~\cite{mupots,3dpw,unite}, one can argue that the accuracy of the ground truth is not sufficient or that the variety in terms of scenes, subjects, and backgrounds is not large enough to properly benchmark state-of-the-art methods. 
Therefore, it is difficult to understand what the current state of 3D human pose estimation is, in particular in-the-wild, and what challenges still need to be addressed.

\begin{figure}[h]
\centering
\includegraphics[width=.7\linewidth]{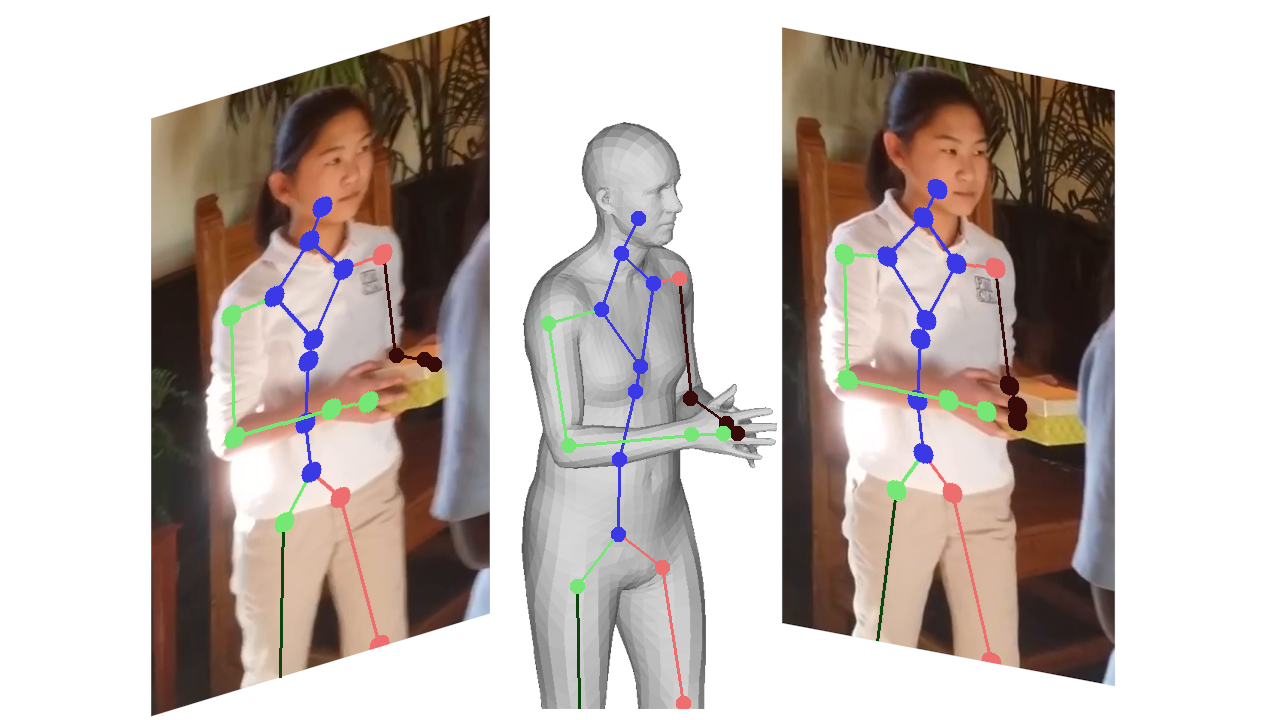} 
\caption{ 3D pose from multiview ``frozen'' people in-the-wild. We build 3D keypoints locations as well as visibility information (occluded joints in dark) with a dense SMPL based approach, based on both 2D and 3D information. Additional examples from various points of view can be found in the supplementary material.}
\label{fig:banner}
\end{figure}

To answer these questions, we present a new dataset of images captured in-the-wild associated with accurate ground-truth body poses.
Some approaches were proposed to annotate images with the SMPL model constraining its optimization with 2D pose detections combined with manual curation~\cite{unite}, Inertial Measurement Units (IMUs)~\cite{3dpw} or using multiple views~\cite{HuangBLKGRAB17}.
One way to easily obtain many views of a person in a certain in-the-wild pose is to capture a video turning around that person while he/she stays still. 
The Mannequin Challenge dataset~\cite{mannequinchallenge}, recently introduced  to predict the depth of humans in an image, shows in-the-wild videos of people, often several persons, frozen in action like mannequins and captured from a moving camera. 
We propose to use this dataset and leverage the static posture of the people with the motion of the camera to apply structure-from-motion techniques and accurately fit the SMPL~\cite{smpl} model on the resulting point cloud. 
More concretely, our pipeline includes the following steps: 
(a) the human instances are tracked and segmented over the frames to find individual instances,
(b) we run a structure-from-motion reconstruction to obtain a 3D point cloud and clean it using the human segmentations,
(c) we use DensePose~\cite{densepose} to obtain correspondences between pixels and SMPL surface, and fit SMPL with the constraint of the pose being constant over frames while minimizing the DensePose reconstruction error. 
After curating the resulting data and checking the quality of our body model registrations with both automatic and manual procedures, we selected for our dataset a total of 742 human subjects in 567 different scenes. We generate in total 24,428 crops around the detected subjects at each frame.
To validate the quality of our reconstruction pipeline, we captured a set of RGB-D sequences and compared the reconstructions produced by our pipeline to the depth data, obtaining a negligible difference.

As a second contribution, we benchmark recent methods that predict SMPL parameters in images~\cite{hmr,nbf,spin,texturepose,graphcmr,dct} or videos~\cite{vibe}. 
We find that the best method regarding 3D pose estimation is VIBE~\cite{vibe} which benefits from leveraging temporal information compared to the other ones that predict SMPL parameters from a static image.
Interestingly, when evaluating the 2D reprojection error of the body joints, HMR~\cite{hmr} outperforms more recent approaches. This can be explained by the fact that HMR was trained on more in-the-wild datasets with 2D annotations.
Despite impressive results, all state-of-the-art methods tend to catastrophically fail when humans are either occluded by objects in the scene in close-up or truncated by image boundaries. The presented benchmark helps understanding such failure cases and we believe it is of great importance to improve robustness of future works.

 
After reviewing related benchmarks in Section~\ref{sec:related}, we present our method to reconstruct static people in Section~\ref{sec:reconstruct} and validate it in Section~\ref{sec:validation}.
Section~\ref{sec:dataset} then summarizes our SMPL Mannequin Benchmark dataset while we benchmark state-of-the-art methods in Section~\ref{sec:results}.

\section{Related Benchmarks}
\label{sec:related}

\begin{table}
 \centering
\caption{Comparison with recent datasets used for evaluation.}
\resizebox{\linewidth}{!}{
\setlength{\tabcolsep}{2pt}
\begin{tabular}{l|c|c|c|c|c|c|c}
Dataset      & \makecell{\#\\Frames}         &  \makecell{\#\\Scenes}      & \makecell{\#\\Subjects} & \makecell{In-\\the-\\wild} & \makecell{Max \#\\ subjects \\ per frame} & 3D GT &  GT Source \\
  \hline
  Human3.6M~\cite{h36} & 3.6M & 1 & 2 & - & 1&  keypoints& marker-based \\ 
    \hline
    Panoptic~\cite{panoptic} & 1.5M & 1 & ~40 & - & 8 &  keypoints& multi-views \\
    MARCOnI~\cite{marconi}& 6190&  7& ~10 & \checkmark  & 2 &  keypoints &multi-views \\
MuPoTS~\cite{mupots}&8000 & 20 &  3 & \checkmark  & 3 & keypoints &multi-views \\
\hline
3DPW~\cite{3dpw}        &   51,000        &      60      &  7    & \checkmark    & 2   & SMPL& 1view + IMUs\\
UP-3D~\cite{unite}         &  8515     &   8515   &   8515  & \checkmark  &1  & SMPL & 1view + annot.\\
  \hline
\textbf{Ours}                   & 24,428 & 567 & 742 & \checkmark & 5 & SMPL & video + static \\

 \end{tabular}
}

 \label{tab:datasets}
\end{table}

Several datasets have been employed to evaluate 3D human pose estimation methods. 
These include datasets captured in controlled environments such as HumanEva~\cite{humaneva} or Human3.6M~\cite{h36}, semi-synthetic datasets such as MPI-INF-3DHP~\cite{mono-3dhp2017} or MuCo-3DHP~\cite{mupots} and, finally, in-the-wild datasets like MARCOnI~\cite{marconi}, MuPoTS~\cite{mupots}, UP-3D~\cite{unite}, or 3DPW~\cite{3dpw}. 

Datasets captured in controlled environments can rely on marker-based MoCap systems to obtain very accurate ground-truth information~\cite{humaneva,h36}.
Another way to produce ground-truth data is to employ a marker-less multiview MoCap system~\cite{panoptic}.
This technique can also be employed outdoors as in~\cite{mupots, marconi} but this implies setting-up multiple cameras in the scene of interest while making sure that the subjects are fully visible in several views.
In practice, this limits the possible scenes where data can be captured and the quality of the ground-truth 3D poses highly depends on the number of cameras deployed.

All these datasets provide ground-truth for body 3D keypoints. Closer to ours are datasets that take a step further and provide the full 3D shape of the persons. 
This is the case of  3DPW~\cite{3dpw} where an optimization pipeline was employed to fit the SMPL model using 2D pose detections associated to motion data coming from IMUs attached to the persons.
However, the known limitations of IMUs (set-up, initial alignment, accumulated errors) make the process hardly scalable.
Another example is UP-3D~\cite{unite} where the SMPL model was fitted to the images using single-view 2D pose detections and results were manually curated. 
In our case, we also optimize the parameters of the SMPL model but we use multiple views constraints of the persons in the same poses ensuring an accurate 3D pose.
Our pipeline allows to generate ground-truth for partially occluded people and multi-person scenes. Importantly,  it only requires a single camera to generate the data, making our method easily scalable and allowing us to build a more varied dataset with more subjects and scenes compared to other in-the-wild datasets, as indicated in Table~\ref{tab:datasets}.





\section{Reconstructing static human poses in videos}
\label{sec:reconstruct}

\begin{figure*}[h]
\centering
\includegraphics[width=.9\linewidth]{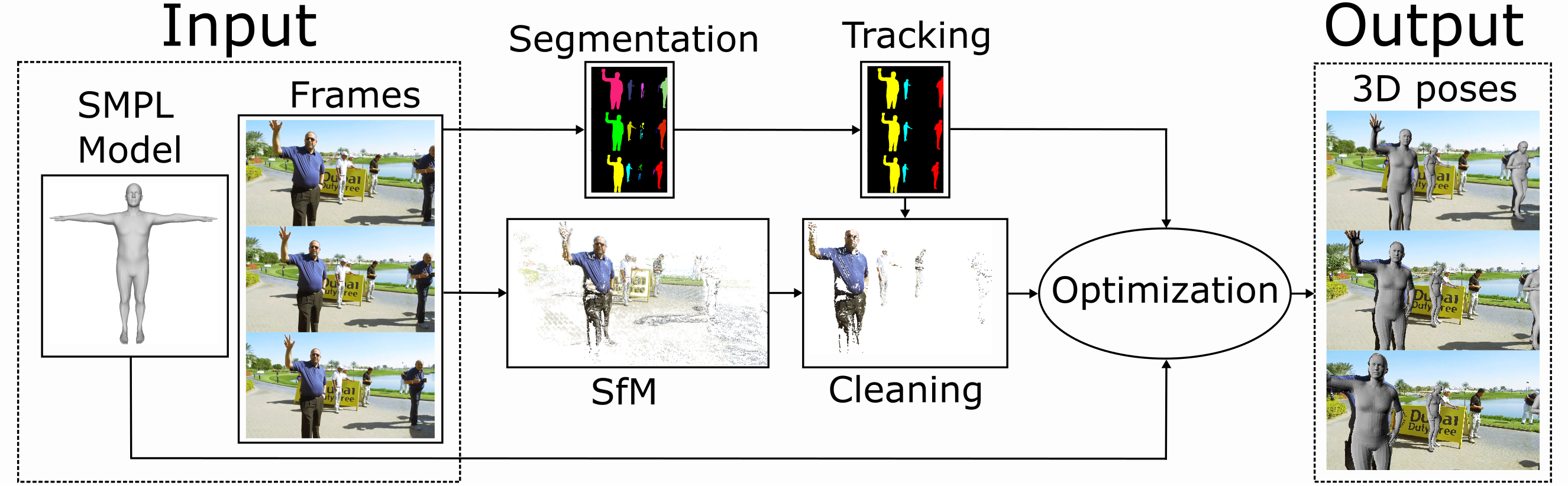} 
\caption{ Our pipeline to reconstruct human pose in-the-wild. See text for details.}
\label{fig:pipeline}
\end{figure*}

To overcome the depth ambiguity inherent to monocular pose estimation methods, we propose to enrich the 3D pose estimation formulation with additional multi-view geometry constraints. In particular, we combine dense constraints arising from the 2D detections in the images in a fashion similar to that in~\cite{kocabas19}, with 3D point clouds reconstructed from the images, like~\cite{jiang19}. 
We build our benchmark upon the Mannequin Challenge dataset~\cite{mannequinchallenge} which consists of video sequences of people mimicking mannequins captured with a single moving camera. 
This section presents the pipeline that we use to robustly reconstruct static human poses in videos, see Figure~\ref{fig:pipeline}.
It consists of three main steps: 
(a) humans are segmented and tracked (Section~\ref{sub:track}), 
(b) the resulting tracks are used to clean the point cloud reconstruction (Section~\ref{sub:clean}), 
(c) SMPL models are fit to each human instance of a sequence with an optimization scheme (Section~\ref{sub:optim}) leveraging 3D information coming from the multi-view scenario.


\subsection{Human instance segmentation and tracking} 
\label{sub:track}

We rely on DensePose~\cite{densepose} to perform human segmentation in each frame, which additionally provides body part segmentation and dense UV coordinates that will be used later. 
For every human instance in an image, we try to associate it with instances in neighboring frames by warping all the pixels of the mask in the central frame to a neighboring frame according to the optical flow (computed with SelfFlow~\cite{selfflow}). 
Because such associations may be noisy, we remove non-maximum associations: at least $k_1=50\%$ of the pixels of the source instance have to fall inside the associated instance in a neighboring frame for the connection to be considered valid. 
Furthermore, we also check forward and backward compatibility: a connection is kept only if it is detected with the forward and the backward pass.
Finally, we obtain all the human tracks by keeping all connected human masks, ignoring clusters with less than $k_2$ nodes, $k_2$ being set to $20 \% $ of the sequence length.
Because the inter-frame motion is rather limited, and thanks to restrictive values for the $k_1$ and $k_2$ thresholds, this tracking method performs well most of the time and discards a track when information is ambiguous. 

\subsection{SfM and Point Cloud Cleaning}
\label{sub:clean}

Our goal is to use 3D knowledge of the scene to better constrain the estimation of the pose, which is an active trend~\cite{jiang19}.
Similar to~\cite{mannequinchallenge}, we perform Structure-from-Motion (SfM) on each sequence using COLMAP~\cite{colmap}. 
These reconstructions contain 3D points of humans, but also on the surrounding environment which needs to be pruned.
To this aim, we count 3D points that are visible and belong to a human according to the segmentations, and only keep them when they appear more often than a given threshold. Additional details can be found in the supplementary material. This validity  is estimated using three following heuristics: \\
\noindent $\bullet$ The 3D point has to reproject inside a human mask. \\
\noindent $\bullet$ It has to be visible,  approximate visibility being computed using a low-resolution soft z-buffer. \\
\noindent $\bullet$ Visibility is reinforced using appearance: the point clouds are equipped with color information from the cameras that agreed on a particular point location, that we use to disambiguate the visibility when the point cloud is incomplete. 

\subsection{SMPL Optimization}
\label{sub:optim}

The segmented point clouds we obtain are often very noisy and incomplete due to motion blur and video compression artifacts.
Thus, we cannot fit SMPL using regular ICP~\cite{bodynet}. 
We consequently devise an optimization scheme that considers per-frame geometric and semantic information.
For every sequence, we are given a set of $N$ images $\{ I_n \}_{n=1}^N$, and their projection operator $\pi_n:  \mathbb{R}^3 \ \to \mathbb{R}^2$. 
For each human instance, we aim at finding the SMPL model that best explains the observed frames in a sequence while accounting for the camera motion. 
More formally, SMPL~\cite{smpl} is a parametric human model that is a function of  pose $\theta$ and  shape $\beta$ parameters. 
%
To find the 3D pose in a sequence, we minimize an objective function (Equation~\ref{eq:obj}) that is the sum of five error terms described in the upcoming paragraphs: a 3D data term accounting for the 3D reconstruction, a 2D reprojection error data term, two pose priors and a shape prior:  

\vspace{-0.3cm}
\begin{equation}
\{ \theta^*,\beta^*\} = \argmin_{\theta,\beta} E_{3D} + E_{2D} + E_{epose} + E_{mpose} + E_{shape}.
\label{eq:obj}
\end{equation}

\noindent \textbf{3D Data term.}
DensePose~\cite{densepose} provides body part segmentations and UV texture coordinates.
The 3D data term ensures that the vertices of the fitted SMPL model roughly match with the extracted UV coordinates in the images.
Let $\mathbf{V}^r$ be the set of $3D$ points in the point clouds after cleaning, \ie, all points are supposed to belong to the human instance. 
For each clean 3D point $x_r \in \mathbf{V}^r$, we compute its pixel projection $\pi_n(x_r)$ in the frame $I_n$ of the sequence.
Given the DensePose pixel-to-vertex association, we recover the corresponding SMPL vertex coordinate $x_s = DP(\pi_n(x_r))$ for this point.
The 3D data term is a cost on the Euclidean distance between SMPL points and target 3D points, summed over all frames and all 3D point clouds:
\vspace{-0.3cm}
\begin{equation}
E_{3D} (\theta,\beta) =  \sum_{n=1}^N \sum_{x_r \in \mathbb{V}^r_n} \omega^{3D}_s \rho( x_r - x_s ),
\label{eq:3Dloss} 
\end{equation} 
with $w^{3D}_s$ which weights the contribution of each point $s$ depending on the number of times it was associated and $\rho$ the Geman-McClure penalty function~\cite{mcclure} that has shown robustness with noisy estimates.

\noindent \textbf{2D Data term.} 
Because the 3D point cloud is not complete and often noisy, we devise a reprojection error loss based on DensePose UV coordinates. 
These coordinates map vertices of the SMPL model to pixels in the image. 
For every pixel in the mask $p \in M$, we project the associated $3D$ SMPL point and penalize the distance in the image plane, for every frame $n$ of the track:

\vspace{-0.3cm}
\begin{equation}
E_{2D} (\theta, \beta) = \sum_{n=1}^N \sum_{p \in M} \omega^{2D}_s \rho (p - \pi_n(DP(p))).
\label{eq:2Dloss}
\end{equation}

Both weights $w^{2D}_s$ and $\omega^{3D}_s$ mean that model points that were respectively not often observed nor associated will only marginally impact the optimization. These weights are respectively normalized by the total number of appearances or associations and both by the number of frames in the track.

The remaining terms are used as defined in \cite{bogo16} and briefly explained for completeness' sake: 

\noindent \textbf{Exponential pose regularization.}
We use a pose prior penalizing elbows and knees that bend unnaturally in the form of an exponential penalty:

\begin{equation}
E_{epose} (\theta) = \sum_i \exp(\theta_i).
\label{eq:exppose}
\end{equation}

\noindent \textbf{Gaussian Mixture Model.} 
We penalize non-plausible poses using the prior of~\cite{cmu}:

\vspace{-0.3cm}
\begin{equation}
E_{mpose} (\theta) = \min_j(-\log(g_j \mathcal{N}(\theta;\mu_{\theta,j},\Sigma_{\theta,j}))),
\label{eq:gmmpose}
\end{equation}
with $g_j$ the mixture model weights. 

\noindent \textbf{Shape.} 
We add a regularization term on possible body shapes based on Principal Component Analysis of the SMPL training set:

\vspace{-0.3cm}
\begin{equation}
E_{shape} (\beta) = \beta^T \Sigma_\beta^{-1}\beta,
\label{eq:shapeloss}
\end{equation}
where $\Sigma_\beta^{-1}$ is a diagonal matrix with the squared singular values estimated via Principal Component Analysis from the shapes in the SMPL training set.


It is worth mentioning that we also tried to constrain the problem using a 2D joints detector (OpenPose \cite{openpose}) and adding another 2D reprojection error term on the joints like ~\cite{kocabas19}. Unfortunately, this did not yield enough robustness in our scenario. Because of the small variations in the  the detected joint locations, the triangulated 3D points often fell far away from the real objective. We illustrate this in the supplementary material by showing that using a denser representation is less prone to errors and overall more robust to noise.

\subsection{Post Processing}
\label{sub:postprocessing}

The obtained poses can be quite noisy in some cases and the optimization often diverges for various reasons. 
Such reasons could be that DensePose UV coordinates are not consistent along a sequence, tracking fails or 3D points are wrongly associated to model vertices. 
To tackle these, we devised three post-processing steps in order to keep only successfully reconstructed humans from the sequences: reprojection  check, visibility check and manual verification.

\noindent \textbf{Reprojection check.} This is a simple automatic reprojection error verification step. 
Estimated poses are kept only if the residual reprojection error is below a restrictive threshold. This excludes most of the erroneous estimated poses of the dataset. 

\noindent \textbf{Visibility Check.} Many humans are not correctly observed in a sequence, \eg only a single limb is visible. In order to discard such examples, we compute joints' sequence-wise visibility: a joint is considered to be visible in the sequence if it reprojects inside a valid body part of DensePose $80$ percent of the time. We only keep models with at least half the torso and one limb visible. The selected models are equipped with this visibility information for evaluation.

\noindent \textbf{Human Verification.} Finally, we perform a manual check.
A human annotator visualizes (a) the 3D SMPL model with its joints visibility along with the 3D reconstruction, and (b) the rendered model on a few images in the track. 
This way, the annotator can discard samples that are inaccurate, either far from the reconstruction, with an invalid reprojection on the images or not corresponding to the reality. 
This inspection takes less than a minute per instance. 
A single annotator can check more than $600$ instances in $7$ hours.

\section{Validation of the method}
\label{sec:validation}

%
%

\begin{figure}

\subfloat{\includegraphics[width=.3\linewidth]{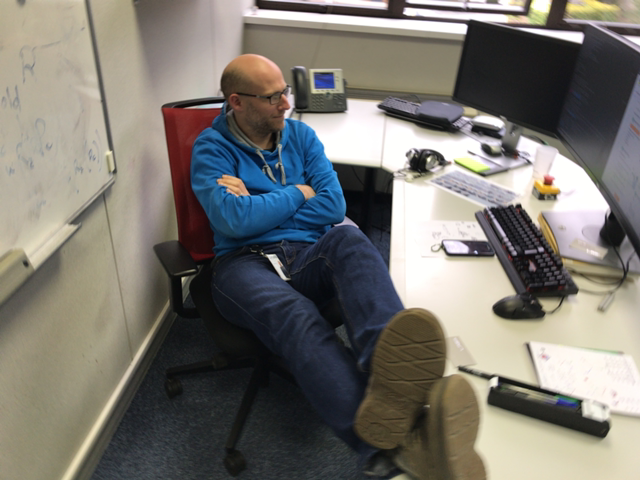}} \hfill
\subfloat{\includegraphics[width=.3\linewidth]{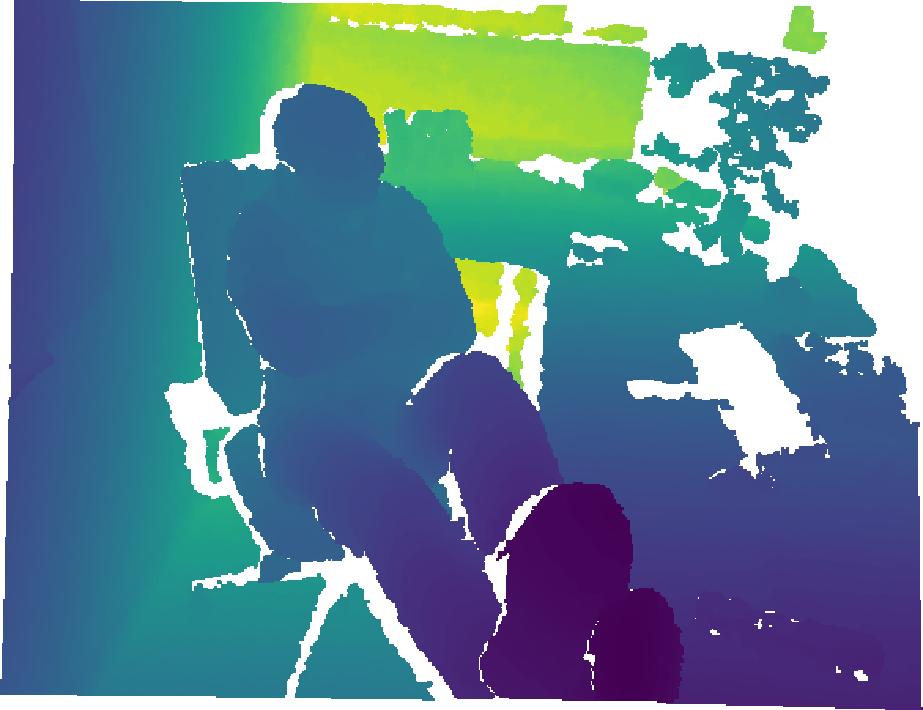}} \hfill
\subfloat{\includegraphics[width=.3\linewidth]{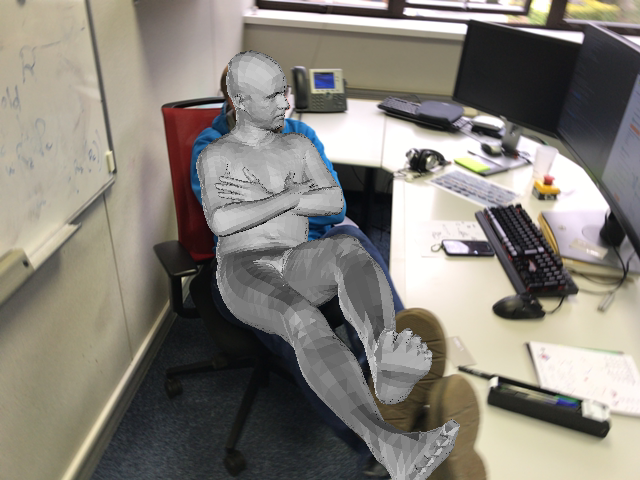}}
\caption{ Example of captured validation sequence. \textit{Left:} RGB image. \textit{Middle:} Associated depth map. \textit{Right:} Estimated pose using RGB information only. More examples can be found in the supplementary material.}
\label{fig:gt_acq}

\end{figure}

To evaluate the accuracy of the proposed pipeline, we captured a dataset using an off-the-shelf Ipad Pro 11~\cite{ipad}, with a Structure depth sensor~\cite{structure} attached to it, providing dense ground-truth 3D measurements of the scene.
Using this dataset, we (a) measure the robustness of our approach, (b) measure the distance between the surface of the reconstructed human poses and the ground-truth point clouds and (c) compare the results of our pipeline with the same optimization scheme applied to the ground-truth point clouds.

\noindent \textbf{Structure Sensor.} The active stereo depth sensor has an accuracy of $1$ percent @ $3.5m$. Depth and RGB acquisitions  are synchronized and aligned using a procedure provided by the constructor. The images and depth maps are saved at $640 \times 480$ px @ $24$ fps.

\noindent \textbf{Acquisitions.} We reproduce acquisition conditions similar to the Mannequin Challenge dataset. We captured immobile humans in indoor scenes using the hand-held device, possibly containing multiple subjects in various poses, with natural noise such as motion blur, occlusions, sensor noise, \etc
We recorded $19$ validation sequences of a few seconds each, resulting in $3,624$ frames for $29$ human instances ($8$ different subjects). 
We then use the RGB images to find the poses with our pipeline, see examples in Figure~\ref{fig:gt_acq}.

\begin{table}

\caption{Validation metrics. First row is our pipeline using COLMAP, second row is replacing the latter with fused ground-truth depth maps in the optimization (values in $cm$).} 

\begin{center}
\begin{tabular}{c|c|c|c|c}
 & SR  &  mean AD  &  median AD  & MPJPE  \\
\hline \hline
RGB & 55\%  & 6.2 & 3.5 & \multirow{2}{*}{3.1}\\\cline{1-4}

Depth & 55\% & 5.0 & 2.2 &  \\
\hline
\end{tabular}
\end{center}

\label{table:valid}
\end{table}

\noindent \textbf{Evaluation.}
Numerical results are reported in Table~\ref{table:valid}.
We first measure the \textit{robustness} of the method as success rate of the reconstruction (SR), \ie, passing all of the post-processing verification checks.
The first row of Table~\ref{table:valid} shows that optimization correctly converges more than half the time when considering RGB sequences. 
We successfully recover $14$ sequences, that is $16$ human instances totaling $2,500$ frames.
Interestingly, if we replace the COLMAP point cloud with the accumulated ground-truth depth maps, the method does not converge either (SR of 55\% for depth also).
This shows that the failures are not due to the accuracy of the SfM reconstruction. 
In practice, convergence mostly fails due to inconsistent human and body part segmentation. 
Human segmentation impacts the tracking, and when the DensePose UV coordinates are not consistent along the sequence, optimal SMPL pose, shape and location are not well defined, leading to a strong divergence during the optimization step. 
We conclude that our pipeline strongly depends on DensePose to provide accurate results.

Next, we measure the \textit{overall accuracy} by averaging over a sequence the mean and median of absolute distances (AD) for each frame between the captured depth maps and a synthetic depth map obtained by rendering the SMPL model in the estimated pose.
 We only focus here on successful pose estimations that were not automatically discarded. 
 We compute the absolute distance only for pixels where depths are defined in both images: because of parallax or hardware limitations, \textit{ground-truth} depth values are undefined for some pixels, as shown in Figure~\ref{fig:gt_acq} (middle column).
Table~\ref{table:valid} shows that we achieve a per-frame mean of $6.2cm$ and a per-frame median of $3.5cm$ on average. 
We additionally report the absolute distances between the depth maps and \textit{reference models}, computed by replacing the COLMAP point cloud with the ground-truth point cloud. 
Because we are explicitly minimizing the metric in our optimization, the \textit{reference models} are naturally closer to the ground-truth. We will use these \textit{reference models} in the last validation step. 

Finally, we measure \textit{pose accuracy} as the distance between joints of the 3D pose of the reconstructed SMPL from RGB and depth data.
This evaluates the impact of the completeness and accuracy of COLMAPs reconstruction on our results. 
We only consider instances where both optimizations converged, and we mask unseen joints using the per-frame visibility (Section~\ref{sub:postprocessing}). 
The average MPJPE as defined in \ref{sub:metrics} over all joints is reported in Table~\ref{table:valid} with $3.1cm$.
This validates that our RGB based pipeline achieves accurate reconstructions according to the depth sensor. 
Considering that the subjects are possibly wearing loose clothing, we would like to emphasis that the shape parameters are only used to better constrain the optimization, but cannot be considered as ground-truth.

\section{The SMPL Mannequin Benchmark}
\label{sec:dataset}

\begin{figure}
\begin{center}
\includegraphics[width=1.\linewidth]{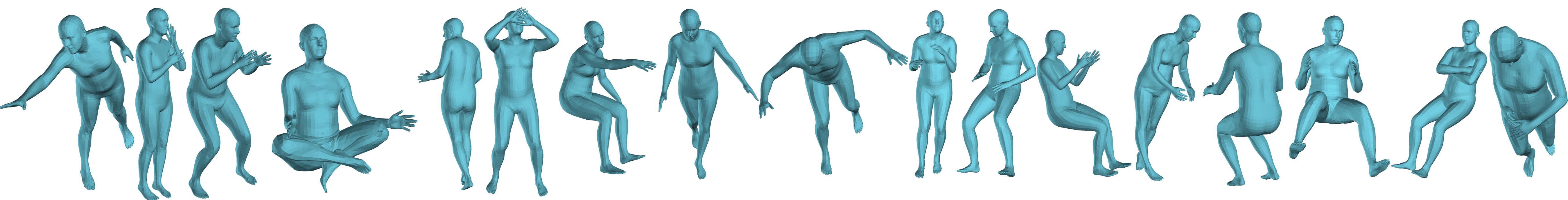} 
\caption{Illustration of pose/viewpoint variability in our SMPL Mannequin Benchmark. }
\label{fig:var}
\end{center}
\end{figure}

In total, we obtained $567$ sequences with $742$ different human instances, resulting in $24,428$ image crops. 
All these crops are equipped with joint visibility information from the post-processing step (sequence-wise and frame-wise), see Section~\ref{sub:postprocessing}.
Our in-the-wild dataset has a strong variability in terms of body poses, appearances and environment, and comprises indoor and outdoor scenes, with natural occlusions and close-ups.
Such natural occlusions with the environment or truncations can be seen in Figure~\ref{fig:hard} (third and fourth columns). 
Moreover, Figure~\ref{fig:var} shows some random poses from our dataset to showcase the variability in poses and viewpoints.

\noindent \textbf{Discussion.}
On one hand, one clear weakness of our pipeline comes from the fact that humans have to stay still during the acquisition. Inherently, we thus cannot capture people jumping or running but only mimes of such motions. 
On the other hand, compared to other in-the-wild 3D datasets~\cite{3dpw,marconi}, we recover the poses of one to two orders of magnitude more different subjects in numerous natural environments, containing occlusions and close-ups, see Table~\ref{tab:datasets}.
Finally, our strategy does not require any particular setup and acquiring new sequences at almost no cost is a matter of seconds with any handheld RGB acquisition device such as smartphones, which are widespread. 
We plan on continuing to increase the size of the dataset by acquiring and processing supplementary sequences. We make it available to the community at the following link: {\small \url{ https://europe.naverlabs.com/research/computer-vision/mannequin-benchmark}}.

\section{Comparison of state-of-the-art methods}
\label{sec:results}

In this section, we benchmark several SMPL-based pose estimation methods.
We describe them in Section~\ref{sub:listmethods} before introducing the metrics used in our benchmark~\ref{sub:metrics}.
Results are presented and discussed in Section~\ref{sub:results}.

\subsection{Evaluated approaches}
\label{sub:listmethods}

We compare the performance of several approaches whose code was available online, namely HMR~\cite{hmr}, NBF~\cite{nbf}, SPIN~\cite{spin}, DCT~\cite{dct}, Texture Pose~\cite{texturepose}, GraphCMR~\cite{graphcmr} and VIBE~\cite{vibe}.
Table~\ref{tab:listmethods} shows an overview of their main features.

In terms of training input, most methods take a single image crop, except NBF that uses a human body part segmentation map, VIBE that processes videos, and TexturePose whose losses are based on consistencies under different frames. 
At test time, they all process a single image crop assuming \textit{a single person}, except VIBE that processes videos.
For VIBE~\cite{vibe}, humans are detected and tracked across frames, and prediction is then made independently on each cropped image based on a per-frame method similar to SPIN, adding an adversarial loss on the SMPL motion. For a fair comparison, we directly give the ground-truth tracks to VIBE and denote the method by VIBE*.

\begin{table}
 \centering
 
  \caption{Overview of evaluated methods.}

 \resizebox{\linewidth}{!}{
   \begin{tabular}{ll|c|c|c|c|c|c|c}

               &                            & HMR          & NBF         & SPIN       & DCT          & Text.Pose    & GraphCMR  & VIBE \\
    \hline   \hline
    \multicolumn{2}{l|}{training input}    & image        & part. seg   & image      & image        & image pair   & image     & video   \\
    \hline
    \multicolumn{2}{l|}{SMPL angles}       & axis-angle   & rot. matrix & 6D rot     & rot. matrix  &  6D rot      & vertices  & 6D rot \\
    \hline
    \parbox[t]{5mm}{\multirow{6}{*}{\rotatebox[origin=c]{90}{losses}}}     &
                 2D joints                 & L2           & L2          &            & L1           & L2            & L1        & L2      \\
               & 3D joints                 & L2           & L2          &            & L2           & L2            &           & L2      \\
               & 3D vertices               &              &             & L2         &              & L2            & L1        &         \\
               & texture                   &              &             &            & L1           & L2            &           &         \\
               & SMPL                      & L2           & L1          & L2         & L2           &              & L2        &         \\
               & adversarial               & $\checkmark$ &             &            &              & $\checkmark$ &           & $\checkmark$ \\
    \hline
 \end{tabular}
 }
 \label{tab:listmethods}

\end{table}

When predicting SMPL parameters, different methods use various representations for the angles in the SMPL pose, either axis-angle~\cite{hmr} (3 dimensions), rotation matrices~\cite{nbf,spin} (9 dimensions) or a 6D rotation representation~\cite{zhou2019continuity}. GraphCMR~\cite{graphcmr} predicts a mesh from which a SMPL model can be fitted while other approaches directly predict SMPL parameters.

In terms of losses, most of them use a loss on the reprojection of the 2D joints (which can be used on any 2D pose estimation dataset like MPII~\cite{mpii} or COCO~\cite{coco}) and on the 3D joints when 3D ground-truth is available like on Human3.6M~\cite{h36} for instance. SPIN~\cite{spin} does not use such losses directly, but the method alternates between training a CNN to predict SMPL parameters and refining these parameters using a variant of SMPLify~\cite{bogo16} which leverages such pose estimation.
Some methods use additional losses on SMPL parameters on data where it is available and/or on vertices from the corresponding mesh.
A few methods have leveraged texture correspondences like in DensePose~\cite{densepose}.
Finally, HMR~\cite{hmr} leverages an adversarial loss to ensure the realism of predicted poses. Such a strategy has been followed by TexturePose.
An adversarial loss is also used in VIBE, but at the video-level to ensure realistic motion of the SMPL model parameters.


\subsection{Metrics}
\label{sub:metrics}

We evaluate metrics for the 3D poses as well as their 2D projections in images.
Since the aim is to predict physical joints of rather large volumes, keypoints for 3D pose have an uncertainty level of a few centimeters, that is a lower bound on the meaningful quantitative evaluation. Regardless of that, the results evidence limits of existing methods \eg showing natural cases where all methods completely fail.

\noindent \textbf{3D poses.}
Various approaches assume different virtual cameras. We thus consider metrics after setting the translational SMPL component to 0.
We extract the 3D coordinates of the 24 joints proposed in the SMPL definition and measure for each joint the average error over all instances where this joint is visible.
We finally report the mean per-joint position error (MPJPE) in millimeters by averaging these values over all joints.
MPJPE has the downside of penalizing methods with outliers. We thus also measure the \textit{PCK3D@X} (percentage of correct keypoints in 3D with a threshold of X mm) for every joint, \ie, the percentages of cases where the joint is predicted with an error below X mm, and average this over all joints. To obtain a single numerical value, we report the AUC (Area Under the Curve) when plotting the PCK3D at various thresholds. In practice, we average the PCK3D@X for X varying from 1 to 500 mm with a step of 1 mm.

\noindent \textbf{2D poses.}
We project the joints using the virtual cameras of each method into the images and get pixel coordinates for each joint.
We also project the ground-truth model and compute the error in pixels, which we normalize by dividing by a scale factor, to achieve invariance \wrt camera distance to the subject.
This scale is obtained by drawing the radius of the sphere centered in joint 0 of the ground-truth of the size of the first bone, which measures roughly 11cm in 3D.
We finally report the normalized error the MPJPE, PCK2D@X with X between 0.05 and 2.00 with a step 0.05 and the AUC.

\subsection{Results and discussion}
\label{sub:results}

\begin{table}

 \caption{Comparison of state-of-the-art methods using the MPJPE metric in 3D (top) and in 2D (bottom) for different sets of joints, as well as average MPJPE over the whole body (\textbf{mean}). The column N shows the average number of such visible sets of joints in the dataset. `PA' refers to results obtained after a Procrustes alignment with the ground-truth. For each row, the best method is shown in bold, while the second best is underlined.}

 \vspace{1mm}
 \centering
 \resizebox{\linewidth}{!}{
 \setlength{\tabcolsep}{1pt}
 \begin{tabular}{llrr|c|c|c|c|c|c|c|c}
  &                      & (njts) &   N   & MEAN      & HMR       & NBF       & SPIN      & DCT       & Text.Pose & GraphCMR  & VIBE*     \\ 
    \hline 
    \hline
 
 \parbox[t]{5mm}{\multirow{9}{*}{\rotatebox[origin=c]{90}{3D (mm)}}} & 
   elbows               & (2)    & 16290 &     307.5 &     208.0 &     406.0 &     \underline{171.0} &     179.3 &     202.6 &     176.9 &     \textbf{158.4} \\ 
 & wrists               & (2)    & 14809 &     493.9 &     227.1 &     516.0 &     \underline{210.7} &     217.9 &     246.4 &     225.1 &     \textbf{203.4} \\ 
 & hands                & (2)    & 13118 &     582.3 &     246.7 &     576.5 &     \underline{238.9} &     240.6 &     275.1 &     257.6 &     \textbf{233.7} \\ 
 & knees                & (2)    &  8521 &     289.6 &     165.9 &     376.8 &     \underline{161.1} &     180.8 &     209.9 &     161.2 &     \textbf{150.9} \\ 
 & ankles               & (2)    &  5135 &     313.5 &     225.2 &     549.9 &     \underline{208.3} &     220.6 &     234.5 &     222.1 &     \textbf{186.4} \\ 
 & toes                 & (2)    &  3428 &     354.9 &     225.7 &     614.7 &     \underline{225.5} &     229.5 &     232.8 &     226.4 &     \textbf{200.7} \\ 
 & neck/head            & (2)    & 22971 &     151.0 &     198.2 &     480.4 &     162.4 &     \underline{156.5} &     176.6 &     179.1 &     \textbf{140.9} \\ 
 & torso                & (10)   & 20112 &     100.4 &     123.8 &     238.0 &      \underline{92.5} &     102.1 &     101.0 &      95.9 &      \textbf{81.3} \\
 \cline{2-12}
 & \textbf{mean}        & (24)   & 15402 &     249.5 &     176.3 &     392.5 &     \underline{153.4} &     161.3 &     173.6 &     160.6 &     \textbf{140.1} \\ 
 \cline{2-12}
  & \textbf{mean} (PA) & (24)& 15402 &     141.2 &     110.7 &     212.5 &      \underline{84.8} &     101.4 &     104.6 &      98.7 &     \textbf{84.5} \\
 
 \hline
 \hline 
  
 \parbox[t]{5mm}{\multirow{8}{*}{\rotatebox[origin=c]{90}{2D (\%)}}} &    
   elbows               & (2)    & 16290 &     -     & \textbf{72.1} &     271.2 &  87.1 &  \underline{77.0} &     121.4 &      98.6 &      87.6 \\                                               
 & wrists               & (2)    & 14809 &     -     & \textbf{82.9} &     273.8 &      97.4 &     101.1 &     127.3 &     102.2 & \underline{91.3} \\
 & hands                & (2)    & 13118 &     -     & \textbf{99.5} &     292.5 &     114.0 &     125.8 &     135.0 &     118.0 & \underline{108.6} \\                                                      
 & knees                & (2)    &  8521 &     -     & \textbf{71.6} &     386.6 &      80.9 &      96.4 &      99.5 &      87.9 & \underline{77.0} \\  
 & ankles               & (2)    &  5135 &     -     & \textbf{87.2} &     470.1 &     114.8 &     104.8 &     101.2 &     125.0 & \underline{88.5} \\
 & toes                 & (2)    &  3428 &     -     & \underline{83.6} &     572.5 &     107.7 &      98.5 &     131.3 &     122.3 & \textbf{76.5} \\                                                      
 & neck/head            & (2)    & 22971 &     -     & \textbf{40.6} &     243.3 &      66.9 &  \underline{46.2} &      79.5 &      92.0 &      55.0 \\                                                      
 & torso                & (10)   & 20112 &     -     & \textbf{57.4} &     267.8 &      69.3 &  \underline{62.3} &     102.0 &      93.8 &      71.1 \\
 \cline{2-12}
 & \textbf{mean}        & (24)   & 15402 &     -     & \textbf{68.7} &     320.7 &      84.6 &      80.1 &     108.8 &     101.2 &      \underline{78.4} \\

    \hline
 \end{tabular}
 }

 \label{tab:mpje}
\end{table}

\begin{figure}
 \centering
 \includegraphics[width=0.45\linewidth]{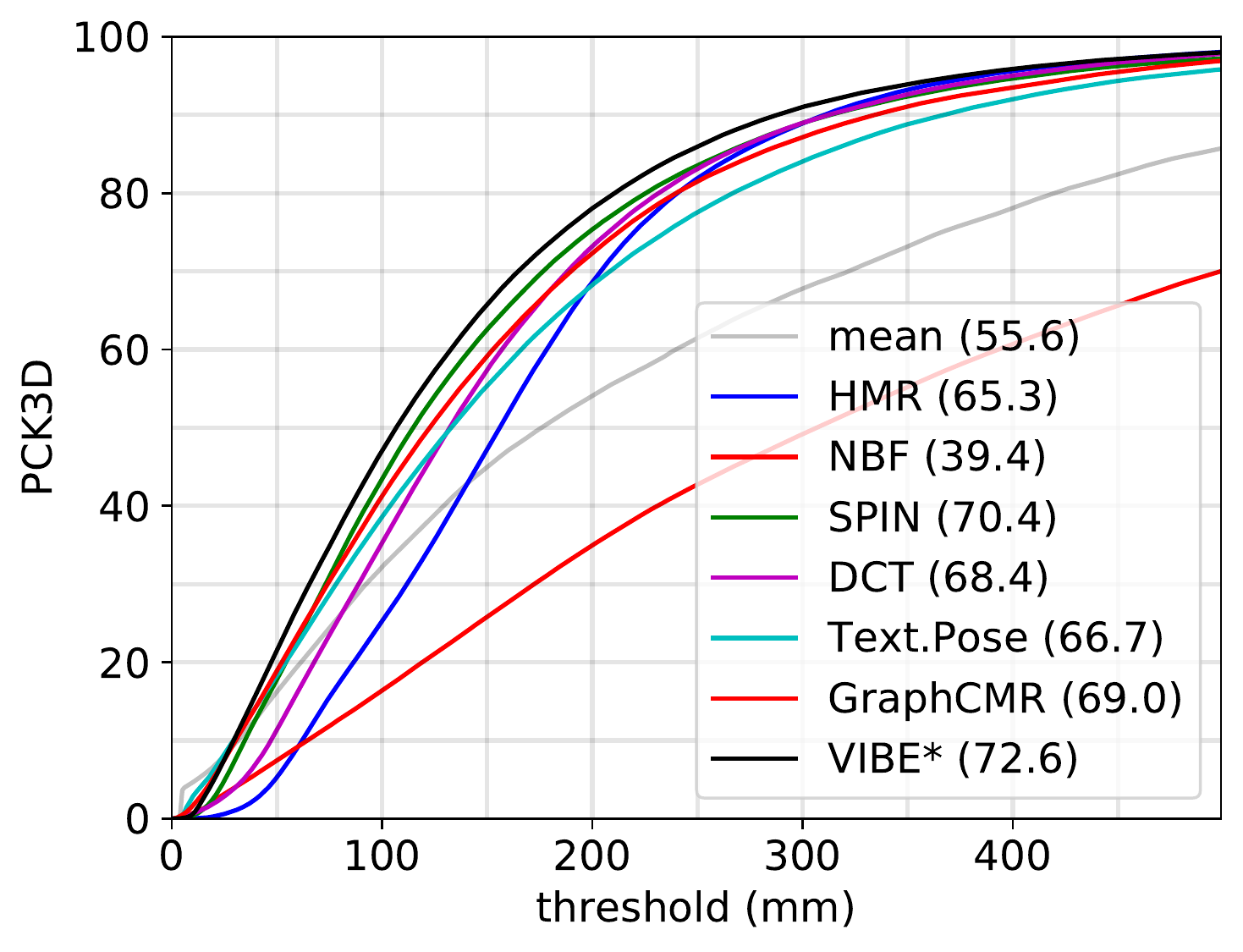} 
 \hspace{0.2cm}
 \includegraphics[width=0.45\linewidth]{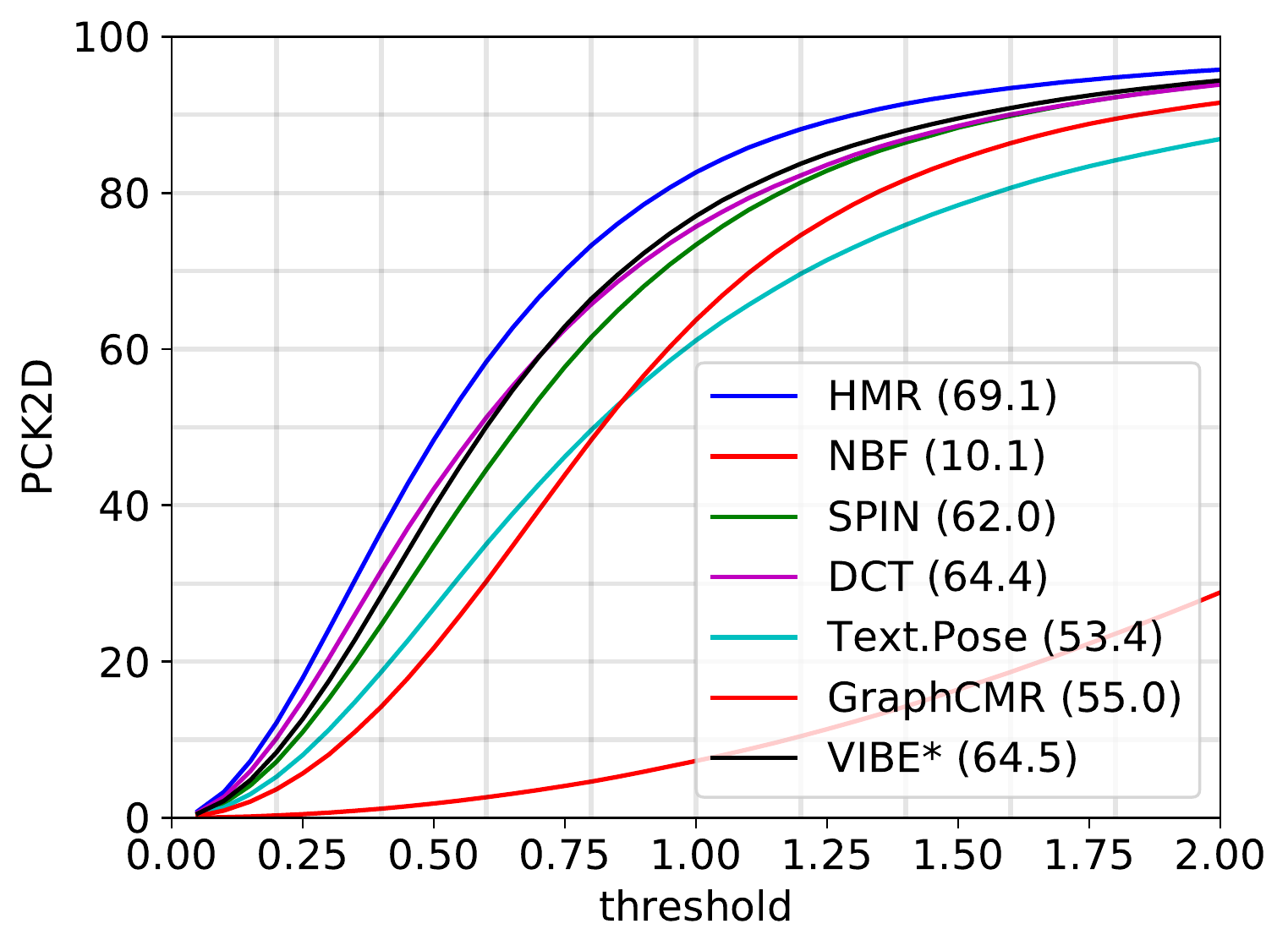}
 \caption{PCK in 3D (left) and in 2D (right) averaged over all joints for varying error thresholds on the SMPL Mannequin Benchmark. AUC is indicated in parenthesis.}
 \label{fig:pck}
\end{figure}

\noindent \textbf{Overall performances.}
We first report the MPJPE in 3D and in 2D in Table~\ref{tab:mpje} for all methods, and for various subsets of joints.
We also report the performance of a naive baseline `MEAN' where the mean pose taken from~\cite{vibe} is returned for any crop.
We additionally plot the PCK3D (left) and PCK2D (right) with their AUC in Figure~\ref{fig:pck}.

Overall, we observe that the farther from the torso the joints are, the higher the error is since their degrees of freedom along the kinematic chain is also higher. 

We observe that NBF~\cite{nbf} performs quite poorly, even worse than the naive MEAN baseline. 
The method predicts SMPL parameters from body part segmentation, and is therefore sensible to this step.
It appears that the body part segmentation fails in many images, resulting in an unrealistic and quite random SMPL estimate.
Among the 12 body parts considered by the method, more than 1,700 images have no segmented part at all, and about 1000 additional ones have fewer than 4 parts, \ie, a total of 20\% of the images, which correspond more or less to the gap in PCK compared to other methods in Figure~\ref{fig:pck} (left).


Overall, VIBE~\cite{vibe} on ground-truth crops (VIBE*) performs best in 3D with 140mm MPJPE and 72.6 AUC.
This is likely because this is the only evaluated method that leverages temporal information.
It actually extends SPIN~\cite{spin} that performs the second best with 153mm MPJPE and 70.4 AUC.
The high performance of this image-based method can be explained by the elegant combination of learning-based and optimization-based approaches during training.
In Table~\ref{tab:mpje}, we also show the MPJPE when applying for each estimated pose a rotation and translation to minimize the MPJPE.
Interestingly, SPIN and VIBE are on par, which means that VIBE likely allows the SPIN estimates to be stabilized over time.

GraphCMR~\cite{graphcmr} and DCT~\cite{dct} follow on the overall ranking and have similar performance.
GraphCMR has a slightly higher AUC (69.0 vs 68.4) with a 1mm lower MPJPE.
By looking at the per joint results, it appears that DCT performs better on farthest joints but worse on joints close to the torso.

\begin{figure}[h]
\begin{center}
\includegraphics[width=\linewidth]{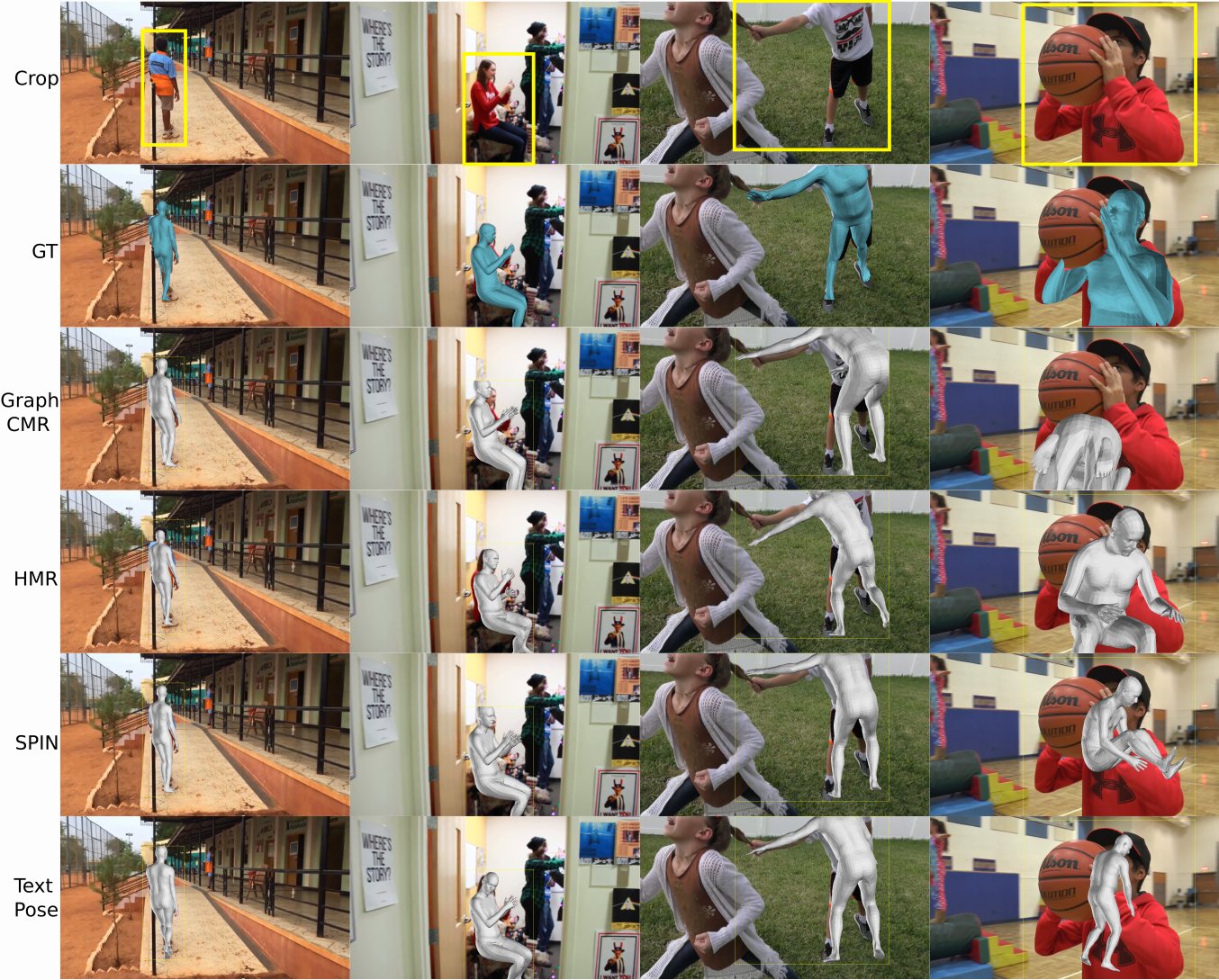} 
\caption{Data samples similarly easy (left) or hard (right) for all methods. See supplementary material for results of all methods.}
\label{fig:hard}
\end{center}
\end{figure}

Next are TexturePose~\cite{texturepose} and HMR~\cite{hmr} with about 175m MPJPE and roughly 66 AUC.
Interestingly, Figure~\ref{fig:pck} (left) shows that HMR reaches this level of AUC with a PCK that is lower at low threshold but higher at high threshold.
In other words, its joint estimates are less aberrant in extreme cases, which may be explained by the use of an adversarial loss.
TexturePose uses the same adversarial loss, and leverages also additional losses like on the consistency of the texture when multiple viewpoints are available.
We explain that it performs on par with HMR by the fact that it uses fewer in-the-wild datasets annotated with 2D pose (MPII only, \vs MPII, LSP, LSPE and COCO for HMR).
Interestingly, this also explains why HMR performs the best in terms of 2D metrics, \ie, after reprojection of the SMPL models into the images.
HMR even outperforms approaches like VIBE or SPIN that achieved better results with regards to the 3D metrics.
The competing approaches using this 2D metric are also trained on varied in-the-wild datasets with 2D pose annotation, \eg UP3D and COCO DensePose for DCT or COCO, MPII and LSPE for SPIN, whereas methods like GraphCMR or TexturePose are trained on fewer in-the-wild training data and perform worse.
In terms of qualitative results, Figure~\ref{fig:hard} displays examples on which methods' performances are correlated, \eg either all performing well ($2$ left columns) or poorly ($2$ right columns). We notice that state-of-the-art methods tend to succeed when the subjects are completely visible with little to no occlusions, but are more prone to errors when truncation or occlusions become stronger. Also, it appears that faces are a dominant discriminating factor, and lead to inverted body orientations or incoherent predictions when hidden (third and fourth columns). 
We evaluate further the impact of occlusions in the next paragraphs.

\begin{figure}
 \centering
 \hspace{-2mm}
  \includegraphics[height=21mm]{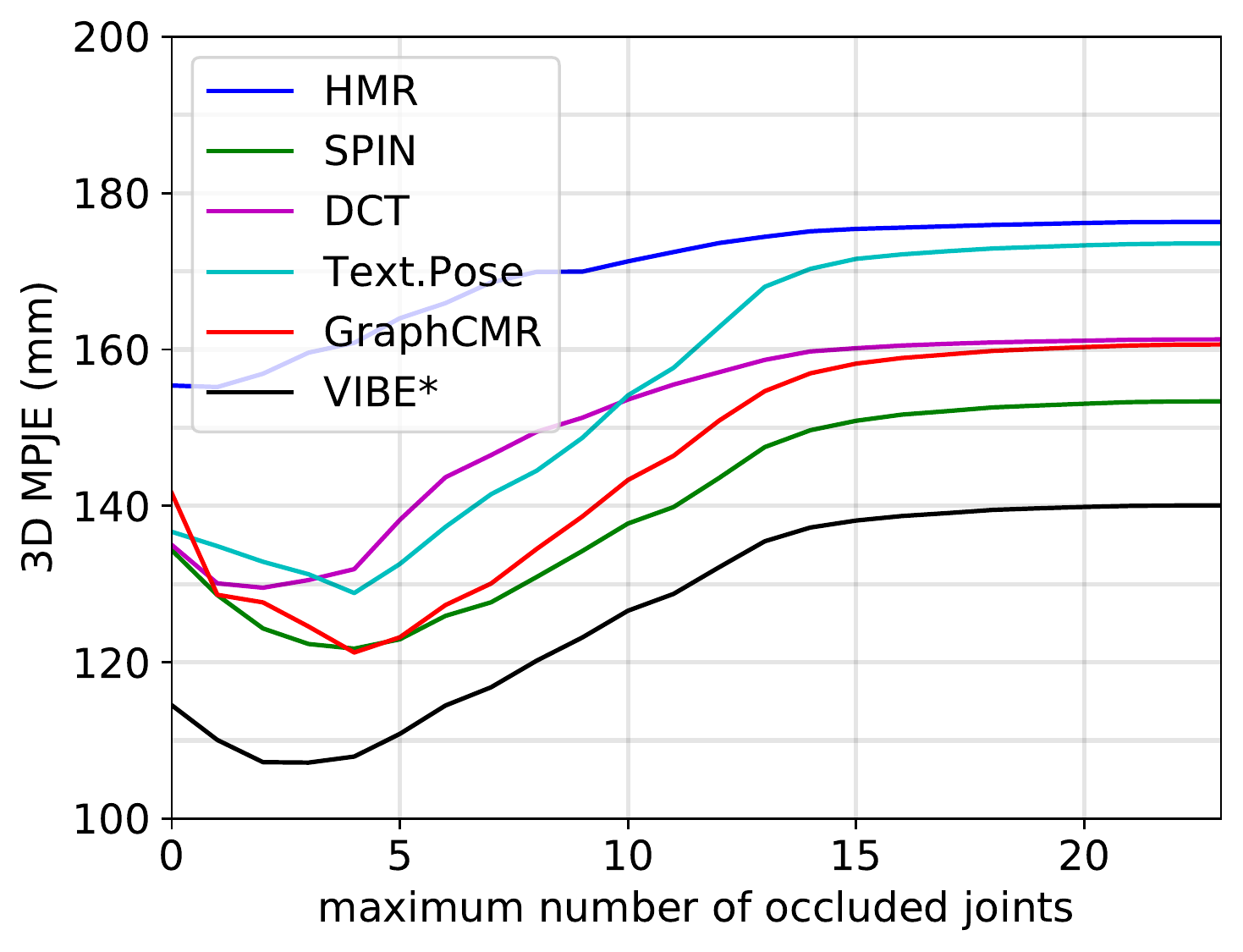}
 \hfill
 \includegraphics[height=21mm]{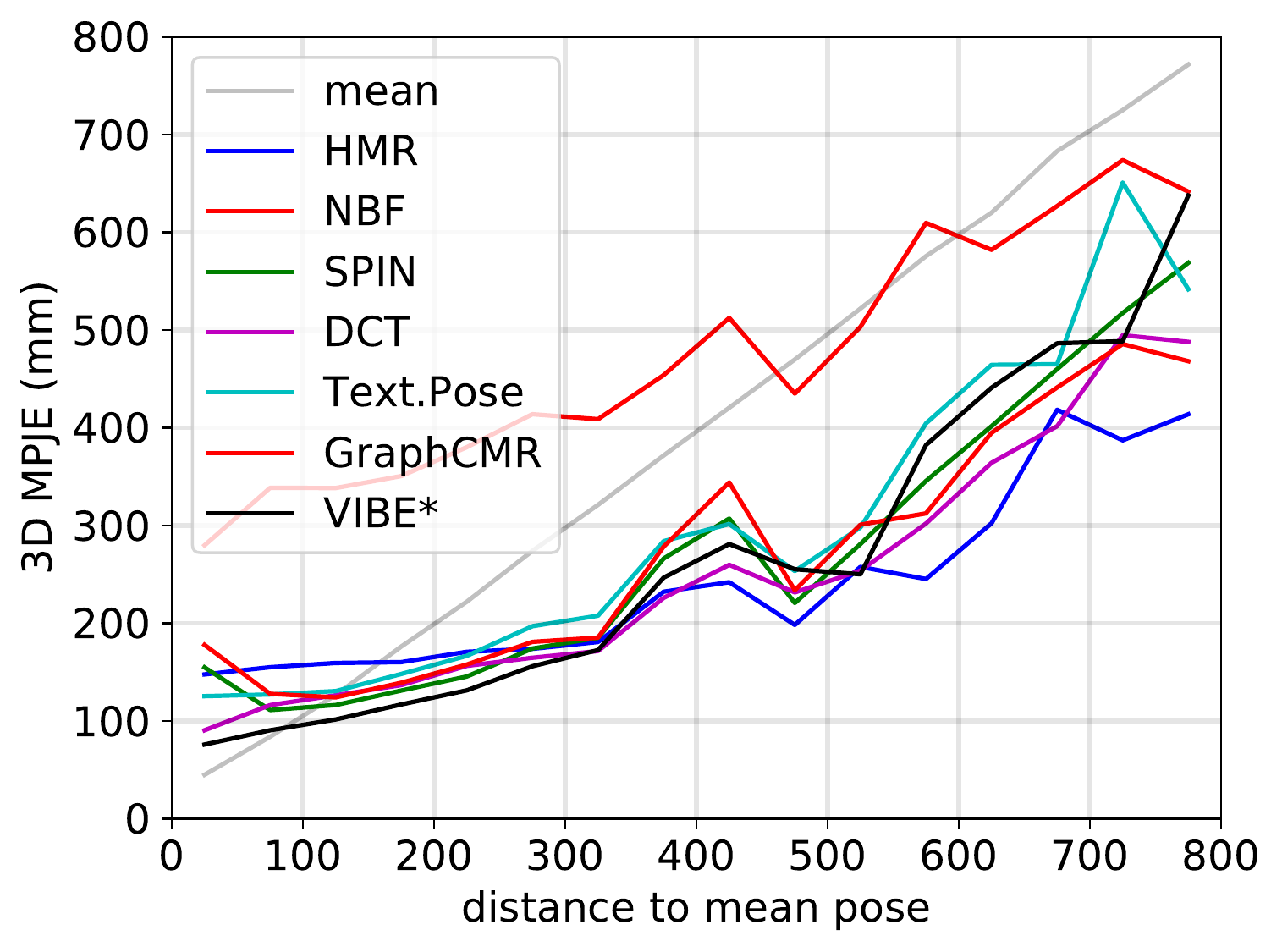} 
 \hfill
 \includegraphics[height=21mm]{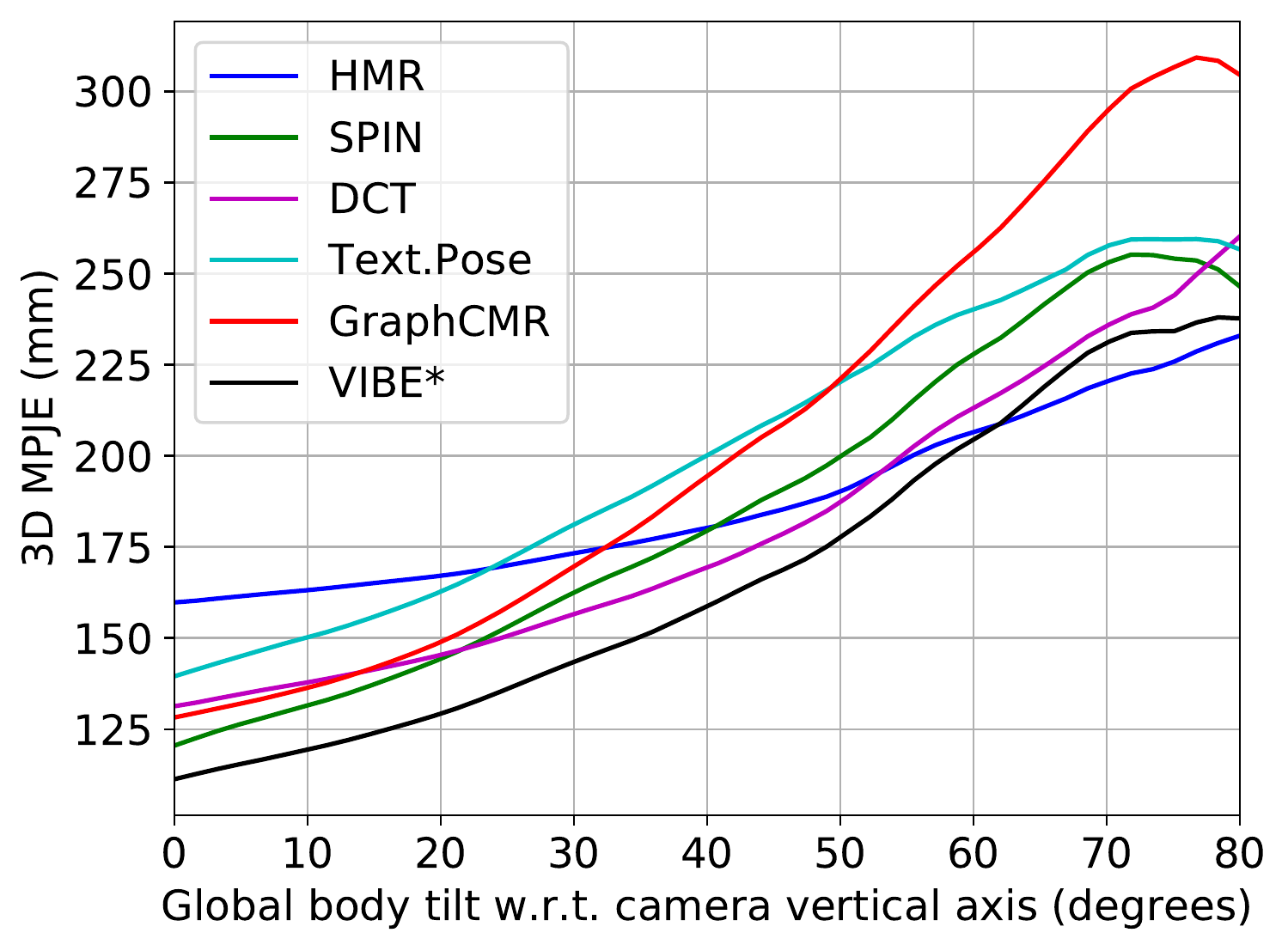}

 \caption{\textit{Left:} 3D MPJPE for the subset of images where the number of occluded joints is bounded by the value on the x-axis. We did not plot NBF whose curve has a similar behavior but starting at 230mm and finishing around 400mm. \textit{Middle:} 3D MPJPE when varying the difficulty of the pose, measured as the distance between the ground-truth pose and the mean pose. The x-axis is based on bins of 50mm. \textit{Right:} 3D MPJPE when varying the tilt angle between the body and the camera vertical axis. We did not plot NBF whose MPJPE increases similarly from 330mm to 500mm.}
 \label{fig:difficulty}
\end{figure}

\noindent \textbf{Impact of occlusions.}
Our dataset contains people with varying  levels of occlusion.
In Figure~\ref{fig:difficulty} (left), we measure the MPJPE of all methods while varying the maximum number of occluded joints.
In other words, when the x-axis has a value of 5, this means that they are maximum 5 non-visible joints, \ie, at least 19 visible joints.
We observe that overall, the MPJPE gets worse when more joints are occluded, and in particular when reaching 10 or 15 occluded joints and more, which means that only a small part of the human is visible, \eg half-body occlusions as visible in last column of figure \ref{fig:hard}.
All methods are approximately as robust/sensitive regarding this aspect.

TexturePose~\cite{texturepose} seems to be more sensitive, which we explain by the lower variety in their training data compared to that of other methods with humans fully visible most of the time.
NBF~\cite{nbf} is not shown on the plot, but the MPJPE when all joints are visible is around 225mm and the plateau on the right side is at 400mm.
The method is thus even more sensitive than the others to occlusions, probably due to the fact that there is fewer body part in the segmentation, leading to aberrant output SMPL parameters.

\noindent \textbf{Impact of pose difficulty.}
Next, we investigate how difficult poses impact the performances of the different methods.
To quantify this \emph{difficulty}, we use the difference (MPJPE) between a given ground-truth pose and the mean pose as a proxy.
We build bins according to this measure and plot the performances of the method according to the ground-truth poses in these bins. 
Results are shown in Figure~\ref{fig:difficulty} (middle).
Clearly, poses farther from the mean pose are harder to estimate for all methods.
It appears that HMR performs the best for difficult poses, likely thanks to its adversarial loss at the image level and its variety of in-the-wild data for training.
DCT also seems to be more robust to difficult poses, as the position of hands (see Table~\ref{tab:mpje}) are quite well estimated compared to other approaches, likely thanks to the loss on the texture while training on COCO-DensePose.

Poses where the global orientation of the body with respect to the vertical axis are not common, \eg  someone lying down, and are also quite difficult in general.
We measure the global body tilt with respect to the camera vertical axis and
plot its impact on 3D MPJPE in Figure~\ref{fig:difficulty}, using a 20\textdegree\ Epanechnikov window smoothing.
Clearly, all methods perform worse when tilt angle increases. Interestingly, HMR~\cite{hmr} again shows better generalization capability as it seems less affected than the other methods.

\section{Conclusion}

We have presented a pipeline to reconstruct 3D poses from videos of still humans. 
We used our method to generate a novel in-the-wild benchmark, using the recent Mannequin Challenge dataset which we validated with depth equipped acquisitions. 
The presented in-the-wild dataset comprises one to two orders of magnitude more subjects than previously existing ones, with accurate ground-truth and with high variability in poses, appearances, environments and orientations. 
Our experiments showed where current state of the art succeeds, but most importantly we unveiled exciting areas of improvement of such approaches, namely strong occlusions and difficult poses, which naturally arise in common videos and that were merely observed with previously existing datasets. 
%

{\small
\bibliographystyle{ieee}
\bibliography{biblio}
}

\end{document}